%% file: main.tex
\documentclass[10pt,twocolumn,letterpaper]{article}

\usepackage{iccv}              
\usepackage{tcolorbox}
\usepackage{ragged2e} 

\definecolor{iccvblue}{rgb}{0.21,0.49,0.74}
\usepackage[pagebackref,breaklinks,colorlinks,allcolors=iccvblue]{hyperref}
\usepackage[T1]{fontenc} 
\def\paperID{16} 
\def\confName{ICCV}
\def\confYear{2025}

\title{Few-Shot Vision-Language Reasoning for Satellite Imagery via Verifiable Rewards}

\author{Aybora Köksal \quad \quad \quad A. Aydın Alatan\\
Dept. of Electrical and Electronics Engineering, Center for Image Analysis (OGAM)\\
Middle East Technical University (METU), Ankara, Turkey \\
{\tt\small aybora@metu.edu.tr \quad \quad \quad alatan@metu.edu.tr}
}

\begin{document}
\maketitle
\input{sec/0_abstract}    
\input{sec/1_intro}

\input{sec/2_related}
\input{sec/3_method}
\input{sec/4_experiments}

\input{sec/5_ablation}
\input{sec/6_conclusion}
{
    \small
    \bibliographystyle{ieeenat_fullname}
    \bibliography{main}
}

\end{document}

%% file: sec/0_abstract.tex
\begin{abstract}
Recent advances in large language and vision-language models have enabled strong reasoning capabilities, yet they remain impractical for specialized domains like remote sensing, where annotated data is scarce and expensive. We present the first few-shot reinforcement learning with verifiable reward (RLVR) framework for satellite imagery that eliminates the need for caption supervision--relying solely on lightweight, rule-based binary or IoU-based rewards. Adapting the “1-shot RLVR” paradigm from language models to vision-language models, we employ policy-gradient optimization with as few as one curated example to align model outputs for satellite reasoning tasks. Comprehensive experiments across multiple remote sensing benchmarks--including classification, visual question answering, and grounding--show that even a single example yields substantial improvements over the base model. Scaling to 128 examples matches or exceeds models trained on thousands of annotated samples. While the extreme one-shot setting can induce mild, task-specific overfitting, our approach consistently demonstrates robust generalization and efficiency across diverse tasks. Further, we find that prompt design and loss weighting significantly influence training stability and final accuracy. Our method enables cost-effective and data-efficient development of domain-specialist vision-language reasoning models, offering a pragmatic recipe for data-scarce fields: start from a compact VLM, curate a handful of reward-checkable cases, and train via RLVR. Our model, training code and dataset will be at \url{https://github.com/aybora/FewShotReasoning}.
\end{abstract}

%% file: sec/1_intro.tex
\section{Introduction}
\label{sec:intro}

\begin{figure}[htbp]
    \centering
    \includegraphics[width=1\linewidth]{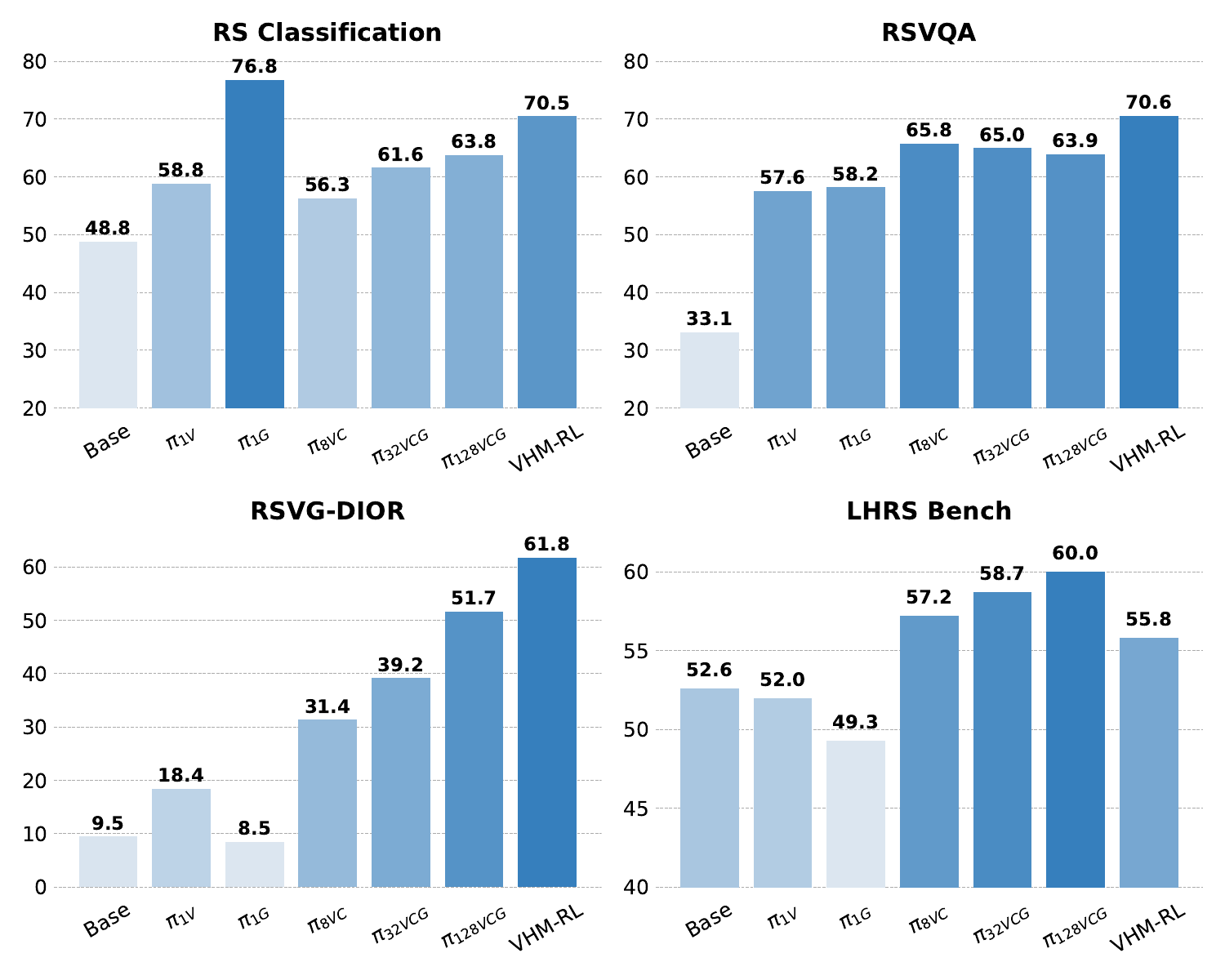}
    \caption{We show that few-shot RLVR achieves strong generalization across remote sensing tasks. With as few as 1–128 reward-checkable examples, our method outperforms or rivals baselines with thousands of examples across RS Classification, RSVQA, RSVG-DIOR, and LHRS Bench, without requiring dense annotations.}
    \label{fig:teaser}
\end{figure}

In last couple of years, advancements in Large Language Models (LLMs) \cite{radford2018improving, Brown2020, achiam2023gpt, touvron2023llama, bai2023qwentechnicalreport, 2023internlm} have unlocked the capabilities of natural language, while the introduction of Vision Language Models (VLMs) \cite{gpt4v, hurst2024gpt, liu2023llava, wang2024qwen2, chen2024internvl} have started a new era in the combination of visual and textual understanding. Recently, Reinforcement Learning with Verifiable Reward (RLVR) methods have been introduced \cite{o1preview, o3mini, guo2025deepseek, team2025kimi} to use reinforcement learning methods to language models with rule based rewards which check whether the final answer is correct or not, or the answer follows a specific format, etc. Yet these models remain general-purpose and computationally heavy, and not that great for specialized domains, such as remote sensing (RS).

As a result, RS-specific VLMs have emerged \cite{hu2307remote, kuckreja2024geochat, zhang2024earthgpt, bazi2024rs, muhtar2024lhrs, pang2025vhm, koksal2025tinyrs}. These developments show the increasing influence of these models in improving both semantic and spatial comprehension for satellite imagery tasks, such as visual question answering, object detection, and grounding in RS scenarios. However, they also introduce a new challenge: requiring hundreds of thousands \cite{kuckreja2024geochat, li2024lhrs} or even million \cite{zhang2024earthgpt, pang2025vhm, koksal2025tinyrs} annotated data to get a proper performance increase without overfitting. Moreover, unlike natural images, satellite scenes demand expert-level annotation: average humans struggle to produce accurate captions for subtle spectral differences or geographic patterns, so they use LLM-generated “pseudo-captions” \cite{muhtar2024lhrs, pang2025vhm}, which often lack the precision needed for fine-tuning.  Manual collection of paired imagery and detailed captions is thus both time-consuming and costly, constraining the size and diversity of available datasets.

A promising alternative emerges from few-shot RLVR: in the context of mathematical reasoning with text-only LLMs, “1-shot RLVR” has demonstrated that training on a single, carefully selected example can achieve performance comparable to training on thousands, by leveraging policy-gradient updates guided by a binary reward signal \cite{wang2025reinforcement}. To our knowledge, this approach has not yet been explored for vision-language models. In the remote sensing domain, RLVR -- let alone few-shot RLVR -- remains largely underexplored, with only a few very recent preprints. \cite{muhtar2025quality} and \cite{koksal2025tinyrs} employ GRPO \cite{shao2024deepseekmath} for alignment following supervised fine-tuning on millions of examples. \cite{koksal2025milchat} attempts to apply RLVR directly without any pretraining as an ablation study; however, it is evaluated only on a small, task-specific dataset, leaving the general-purpose performance on remote sensing tasks still unknown.

Building on recent claims that base language models already possess reasoning capabilities \cite{guo2025deepseek, gandhi2025cognitive, liu2025understanding, yue2025does, shah2025rethinking}, we ask: \textit{Can robust, domain-specific reasoning over satellite imagery be elicited using only a handful of examples and verifiable rewards, rather than full captions?} Following the approach of \cite{wang2025reinforcement}, we use RLVR to train remote sensing reasoning models with training sets as small as a single example. Our main contributions and observations are as follows:

\begin{itemize}
    \item \textbf{A caption-free few-shot RLVR framework for satellite imagery.} We introduce the first vision-language reasoning model for remote sensing trained solely via reinforcement learning with verifiable (binary or IoU-based) rewards--no human- or AI-generated captions required--demonstrating that base VLMs can be aligned for domain-specific reasoning with minimal supervision.
    
    \item \textbf{Unlocking reasoning with as little as one example.} We show that a single curated training example yields double-digit gains over the untouched base model across classification, VQA, and grounding benchmarks, and that performance scales smoothly--matching a 2000-image fully supervised baseline by 128 shots.
    
    \item \textbf{Characterizing and containing overfitting in the 1-shot setting.} Unlike prior text-only RLVR work \cite{wang2025reinforcement}, our 1-shot models exhibit mild, task-confined overfitting (limited to the exact dataset split of the training example) while retaining robust generalization on all other remote-sensing tasks.
\end{itemize}

%% file: sec/2_related.tex
\begin{table}[htbp]\footnotesize
\centering
\caption{A comparative study of our approach with prior work. Vision: Vision-language model, RS Adapted: Adapted to remote sensing problems with domain specific data, RL Reasoning: Trained as reasoning model with an RL based method, Few Shot: Trained with a few examples.}
\scalebox{0.9}{
\begin{tabular}{c|ccccc} \hline
     Method & Year & Vision & RS Adapted & Reasoning & Few Shot \\ \hline
     RSGPT \cite{bazi2024rs} & 2023 & \checkmark & \checkmark & $\times$ & $\times$ \\
     GPT-4o \cite{hurst2024gpt} & 2024 & \checkmark & $\times$ & $\times$ & $\times$ \\ 
     Qwen2-VL \cite{wang2024qwen2} &  2024 & \checkmark & $\times$ & $\times$ & $\times$  \\
     EarthGPT \cite{zhang2024earthgpt} & 2024 & \checkmark & \checkmark & $\times$ & $\times$ \\
     GeoChat \cite{kuckreja2024geochat} & 2024 & \checkmark & \checkmark & $\times$ & $\times$ \\ 
     LHRSBot \cite{muhtar2024lhrs} & 2024 & \checkmark & \checkmark & $\times$ & $\times$ \\ 
     o1-o4 \cite{o1preview} & 2025 & \checkmark & $\times$ & \checkmark & $\times$ \\
     TinyRS \cite{koksal2025tinyrs} & 2025 & \checkmark & \checkmark & \checkmark & $\times$ \\ 
     Wang et. al. \cite{wang2025reinforcement} & 2025 & $\times$ & $\times$ & \checkmark & \checkmark \\
     \hline
     Ours  & & \checkmark & \checkmark & \checkmark & \checkmark \\ \hline
\end{tabular}}
\label{table:comparative_summary}
\end{table}

\section{Related Work}
\label{sec:related}

A comparative summary of how our study differs from the related work can be found in Table \ref{table:comparative_summary}.

\noindent \textbf{Language Models.} The introduction of the Transformer architecture \cite{Vaswani2017} ignited a revolution in large language models, first through BERT \cite{Devlin2019} and GPT \cite{radford2018improving}, then culminating in milestone models like GPT-3 \cite{Brown2020}. The arrival of LLaMA \cite{touvron2023llama} flipped expectations by demonstrating that smaller models could actually surpass their heavyweight predecessors, fueling a surge of open-source LLMs such as Qwen \cite{bai2023qwentechnicalreport}, InternLM \cite{2023internlm}, and Gemma \cite{gemmateam2024gemmaopenmodelsbased}. In parallel, CLIP \cite{radford2021learning} set off the era of multimodal models, paving the way for vision-language giants like GPT-4V \cite{gpt4v}, Qwen-VL \cite{bai2023qwen}, and PaliGemma \cite{beyer2024paligemma}. As this landscape matured, it gave rise to compact, highly capable sub-7B multimodal SLMs--most notably, PaliGemma 2 \cite{steiner2024paligemma} and Qwen2-VL-2B \cite{wang2024qwen2}. All of these language models have been a breakthrough, either because of they set a new benchmark, or because of their efficiency in size. However, none of them are originally designed for being a reasoning model. On the other hand, several works show that they are naturally designed for being a task-specific reasoning model \cite{gandhi2025cognitive, liu2025understanding, yue2025does, shah2025rethinking}. Also it is known that with thoughtful adaptation, smaller and more specialized models are now outperforming their larger, generic ancestors in both speed and task-specific accuracy \cite{bucher2024fine, gpt35turbo}.

\noindent \textbf{Reinforcement Learning with Verifiable Rewards.} The popularity of RLVR on the Large Language Models has started with the introduction of OpenAI's o-series models \cite{o1preview, o3mini}. DeepSeek-R1 from DeepSeek \cite{guo2025deepseek} broke new ground in reasoning by sidestepping traditional supervised fine-tuning. Its sibling, DeepSeek-R1-Zero, leverages massive-scale reinforcement learning to self-improve reasoning abilities, but struggles with language mixing. To address this, DeepSeek-R1 introduces a clever blend of cold-start data and multi-stage training, enabling it to rival OpenAI-o1-1217 in performance. By adopting Group Relative Policy Optimization (GRPO) \cite{shao2024deepseekmath}, DeepSeek-R1 delivers strong mathematical reasoning with impressive efficiency. Knowledge distillation further extends its impact, shrinking its powerful reasoning into smaller models (spanning 1.5B to 70B parameters) that outperform prior generations.

Still, while these advanced models leap ahead of vanilla language models in reasoning, they remain broad generalists and miss the benefits of task-tailored fine-tuning. Unlike OpenAI’s proprietary o-series, which supports multimodal tasks, DeepSeek-R1’s focus is unimodal. Yet GRPO opens the door for the open-source community: it provides a blueprint for transforming any LLM or VLM into a focused reasoning model \cite{openr1, openr1multimodal, openr1v}, paving the way for small, efficient, task-specific, and even multimodal open-source reasoning systems.

\noindent \textbf{Few Shot RLVR.} Wang et al. \cite{wang2025reinforcement} show that reinforcement learning with verifiable reward (RLVR) can dramatically improve large language model reasoning with as little as one training example (“1-shot RLVR”). Training Qwen2.5-Math-1.5B on a single example raises its MATH500 accuracy from 36\% to 73.6\%, matching results from much larger datasets. The study finds that this improvement holds across models and tasks, and uncovers effects like post-saturation generalization, cross-domain gains, and increased self-reflection. The work suggests that strong reasoning capabilities already exist in base LLMs and can be efficiently unlocked with minimal, carefully chosen data and RL-based fine-tuning. However, as this study is limited to single-modality language models, it remains unclear whether few-shot RL training of vision-language models (VLMs) on task-specific examples--such as those in remote sensing--would yield similar improvements.

\noindent \textbf{Remote Sensing VLMs.} Recent advances in vision language models (VLMs) have been extended to remote sensing applications. Efforts such as RSGPT \cite{hu2307remote} and GeoChat \cite{kuckreja2024geochat} have introduced new benchmarks (RSICap, RSIEval) and large instruction datasets, using LoRA-based fine-tuning \cite{Hu2022} for tasks like captioning, visual question answering, and spatial interaction. EarthGPT \cite{zhang2024earthgpt} integrated different input types--including optical, SAR, and infrared imagery--into a single framework, while RS-LLaVA \cite{bazi2024rs} improved multi-task performance through LoRA. LHRS-Bot \cite{muhtar2024lhrs} and LHRS-Bot-Nova \cite{li2024lhrs} focused on geographic reasoning using datasets such as LHRS1-Align and LHRS-Instruct. The VHM model \cite{pang2025vhm} introduced a large remote sensing pretraining set with more than 1 million images and 178,000 instructions. Despite these developments, none of these models have applied reinforcement learning (RL) for reasoning in remote sensing tasks. Recently, we found that a couple of preprints tries to do that.

Muhtar et. al. \cite{muhtar2025quality} introduce the 7B ScoreRS model for answer quality analysis, employing Qwen2-VL-RS and Qwen2-VL-RS-R1 with GRPO-based rewards. Similarly, Koksal et. al. \cite{koksal2025tinyrs} claims that higher gains are achieved than Qwen2-VL-RS-R1 even though they are using a smaller model. However, both of these approaches have multi-stage training approaches including supervised fine tuning stages using generated captions from LLMs for millions of images. On the other hand, MilChat \cite{koksal2025milchat} applies RL to military base classification with strong results, while their final work focuses on multi-stage training, they have also completed a study with only RL training. However, their results are very limited to a single task and their generalist performance cannot be observed. Therefore, the performance of generalist language models with only RLVF training, yet alone a few shot case, remains unknown.

%% file: sec/3_method.tex
\section{Method}
\label{sec:method}

\subsection{Model}
\label{sec:model}

Our approach builds on Qwen2-VL-2B \cite{wang2024qwen2}, adopting both its model architecture and pretrained weights:

\noindent \textbf{Visual Encoder:} Like Qwen2-VL-2B, our method employs a Vision Transformer (ViT) \cite{dosovitskiy2020image} with 675 million parameters as the visual encoder. This component supports naive dynamic resolution handling \cite{dehghani2023patch} throughout training and inference, enabling flexible input by mapping images of different sizes into variable-length visual token sequences.

\noindent \textbf{Language Model:} At its core, our method uses the same language backbone as Qwen2-VL-2B, initializing from the pretrained Qwen-1.5B model weights.

\noindent \textbf{Position-Aware Vision-Language Adapter:} For efficient multimodal processing, we integrate a vision-language adapter modeled after Qwen2-VL-2B. This adapter leverages a single-layer cross-attention mechanism to condense the sequence of visual features down to a fixed size of 256, while retaining positional context via 2D absolute positional encodings. The resulting compressed features are passed to the language model for downstream tasks.

Since \cite{wang2025reinforcement} have already demonstrated the robustness of RLVR across various LLM architectures, we believe our approach is equally adaptable to vision language models. Their findings indicate that the effectiveness of RLVR does not depend on a specific model, supporting the generalizability of the method.

\begin{table*}[bt]
    \vspace{-1em}
\caption{\textbf{Few-Shot RLVR performance (\%) for different domains in Remote Sensing.}
Here for RS Classification, we consider AID (AID), METER-ML (MML), NWPU (NW), SIRI-WHU (SIRI), WHU-RS19 (RS19) and RS Classification Average (Avg.). For RSVQA, we consider HR Comparison (HC), HR Presence (HP), LR Comparison (LC), LR Presence (LP), LR Rural-Urban (LRU) and RSVQA Average (Avg.). For Visual Grounding, we report precision scores for IoU$\geq$0.5 on RSVG-DIOR (DIOR). For General Remote Sensing Knowledge, we present LHRS-Bench (LHRS) scores. Size refers to the number of distinct training examples used, and Step indicates the frozen iteration number that yielded the best result.
}
\label{tab:results}
    \centering
    \scalebox{0.85}{
    \begin{tabular}{c|c|c|c|c|c|c|c|c|c|c|c|c|c|c|c|c}
        \toprule
        \textbf{Dataset} & \textbf{Size} & \textbf{Step} & \multicolumn{6}{|c|}{\textbf{RS Classification}} & \multicolumn{6}{|c|}{\textbf{RSVQA}} & \textbf{Grounding}  & \textbf{Knw.} \\
        \midrule
        & & & \textbf{AID} & \textbf{MML} & \textbf{NW} & \textbf{SIRI} & \textbf{RS19} & \textbf{Avg.} & \textbf{HC} & \textbf{HP} & \textbf{LC} & \textbf{LP} & \textbf{LRU} & \textbf{Avg.} & \textbf{DIOR} & \textbf{LHRS} \\
        \midrule
        Base & 0 & 0 & 53.0 & 42.8 & 43.6 & 42.4 & 62.1 & 48.8 & 32.4 & 44.6 & 30.4 & 42.1 & 16.0 & 33.1 & 9.5 & 52.6\\
        \midrule
        VHM-RL & 2000 & 1300 & \underline{70.6} & \underline{59.1} & \textbf{72.4} & 61.8 & \underline{88.4} & \underline{70.5} & \textbf{72.4} & 49.9 & \textbf{82.1} & \textbf{75.6} & \textbf{73.0} & \textbf{70.6} & \textbf{61.8} & 55.8\\
        \midrule
        $\pi_{1V}$ & 1 & 1300 & 58.8 & 45.8 & 52.7 & 51.3 & 85.6 & 58.8 & 69.6 & 63.6 & 65.3 & 25.7 & 64.0 & 57.6 & 18.4 & 52.0 \\
        $\pi_{1C}$ & 1 & 100 & 55.1 & 21.1 & 49.7 & 54.3 & 67.6 & 49.6 & 48.4 & \textbf{70.4} & 45.4 & 40.0 & 44.0 & 49.6 & 29.2 & 54.3 \\
        $\pi_{1G}$ & 1 & 700 & \textbf{77.9} & \textbf{69.6} & \underline{69.1} & \textbf{69.5} & \textbf{98.1} & \textbf{76.8} & 63.6 & 60.2 & 62.2 & 48.9 & 56.0 & 58.2 & 8.5 & 49.3 \\
        \midrule
        $\pi_{2VC}$ & 2 & 100 & 60.2 & 20.7 & 52.5 & 59.0 & 86.4 & 55.8 & 69.6 & 63.9 & 65.3 & 26.3 & \underline{71.0} & 59.2 & 15.6 & 48.4 \\
        $\pi_{2G}$ & 2 & 100 & 57.3 & 48.5 & 55.5 & \underline{64.7} & 80.1 & 61.2 & 50.5 & 57.4 & 63.7 & 56.3 & 64.0 & 58.4 & 35.4 & 54.6 \\
        \midrule
        $\pi_{4VC}$ & 4 & 100 & 58.0 & 17.5 & 50.2 & 54.8 & 77.6 & 51.6 & 68.9 & \underline{69.8} & 65.3 & 33.8 & 64.0 & 60.4 & 20.6 & 53.6 \\
        $\pi_{4VCG}$ & 4 & 200 & 55.9 & 29.2 & 50.2 & 49.3 & 72.3 & 51.4 & 70.0 & 67.6 & 65.1 & 30.0 & 63.0 & 59.1 & 23.3 & 54.5 \\
        \midrule
        $\pi_{8VC}$ & 8 & 200 & 58.2 & 39.3 & 50.2 & 57.4 & 76.4 & 56.3 & 70.6 & 58.2 & 67.5 & 63.7 & 69.0 & \underline{65.8} & 31.4 & 57.2 \\
        $\pi_{8VCG}$ & 8 & 100 & 60.3 & 28.5 & 51.1 & 56.2 & 80.7 & 55.3 & 68.0 & 41.6 & 63.1 & \underline{74.8} & 62.0 & 61.9 & 29.9 & 56.7 \\
        \midrule
        $\pi_{16VC}$ & 16 & 800 & 58.4 & 39.1 & 49.5 & 58.2 & 81.8 & 57.4 & 68.2 & 64.0 & 69.6 & 50.1 & 69.0 & 64.2 & 31.8 & 58.1 \\
        $\pi_{32VCG}$ & 32 & 400 & 62.9 & 42.3 & 54.9 & 59.3 & 88.5 & 61.6 & 69.6 & 54.9 & 65.2 & 74.4 & 61.0 & 65.0 & 39.2 & \underline{58.7} \\
        $\pi_{64VCG}$ & 64 & 1900 & 62.9 & 45.5 & 55.9 & 59.4 & 82.4 & 61.2 & 70.6 & 53.3 & 67.8 & 73.2 & 64.0 & 65.8 & 47.3 & 57.5 \\
        $\pi_{128VCG}$ & 128 & 1600 & 67.3 & 48.7 & 60.9 & 56.9 & 85.4 & 63.8 & \underline{71.1} & 60.7 & \underline{69.9} & 73.6 & 44.0 & 63.9 & \underline{51.7} & \textbf{60.0} \\
        \bottomrule
    \end{tabular}
    }
\end{table*}

\subsection{RL with Verifiable Rewards}
\label{sec:rl}

In this work, we leverage Group Relative Policy Optimization (GRPO) \cite{shao2024deepseekmath} to refine our policy using relative reward signals calculated over sets of candidate responses. Drawing inspiration from DeepSeek-R1 \cite{guo2025deepseek}, we introduce two complementary reward types:

\noindent \textbf{Format compliance reward:} a binary signal granted whenever the model’s output strictly follows the prescribed structure: \textit{<reasoning>...</reasoning> <answer>...</answer>}, irrespective of the content within those tags.

\noindent \textbf{Task-specific accuracy reward:} Based on the idea from \cite{koksal2025tinyrs}, we have defined two different accuracy rewards:

\begin{itemize}
    \item Closed-ended tasks (e.g. VQA, classification, multiple-choice): a binary signal (1 if the answer is correct, 0 otherwise).
    \item Visual grounding: the Intersection over Union (IoU) between the predicted and true bounding boxes serves as the base reward value. We build on the ablation study presented in \cite{koksal2025tinyrs}, which demonstrates that their proposed quantized IoU-based reward yields more stable and balanced performance across diverse benchmarks. Motivated by these findings, we adopt the same quantized reward function in this work. Specifically, a reward of 1 is assigned for IoU/score values $\geq 0.7$, an exact base reward is given for values between 0.4 and 0.7, and a reward of 0 is assigned for values $< 0.4$. This scheme is consistently applied to both types of tasks.
\end{itemize}

\subsection{Loss}

To train with GRPO, we introduce two different types of loss function in this work:

\noindent \textbf{The Policy Gradient Loss:} This loss encourages the model to favor responses that yield higher rewards by weighting them according to group-normalized advantages. It reinforces solutions that surpass the group average while suppressing those that underperform. In our case, we group two distinct reward functions -- based on \cite{guo2025deepseek} and \cite{koksal2025tinyrs} -- as detailed in Section \ref{sec:rl}.

\noindent \textbf{The KL Divergence Loss:} It is used to preserve overall language quality by comparing the model’s current outputs to those of a reference model, penalizing deviations between them.

Please note that, unlike \cite{wang2025reinforcement}, we did not use entropy loss during training, as they also mention that it is not required, and the DeepSeek paper does not introduce it either.

\subsection{Dataset}
\label{sec:dataset}

For this work, firstly random 2000 images are sampled from existing VHM-Instruct dataset \cite{pang2025vhm} to generate the reference VHM-RL subset. These images are selected from the following tasks:

\begin{itemize}
    \item VQA: Visual Question Answering (520 images)
    \item CLS: Scene Classification (637 images)
    \item VG: Visual Grounding (843 images)
\end{itemize}

\noindent To focus on the task-specific accuracy reward described in Section~\ref{sec:rl}, we sample our few-shot training sets (ranging from 1 to 128 examples) directly from the reference dataset. In contrast to \cite{wang2025reinforcement}, which explored historical variance-based selection, our sampling is entirely random. Although their paper proposes a data selection strategy based on historical variance, the authors acknowledge that this method is not universally optimal and that randomly selected examples can sometimes yield up to 30\% better performance. Therefore, we align with their ultimate conclusion that the observed few-shot RLVR phenomenon is largely independent of the data selection method, and we opt for random sampling in our experiments.

The number of examples from each dataset used in our final models is defined as follows:

\begin{itemize}
    \item $\pi_{1V}$: 1 VQA
    \item $\pi_{1C}$: 1 CLS
    \item $\pi_{1G}$: 1 VG
    \item $\pi_{2VC}$: 1 VQA and 1 CLS
    \item $\pi_{2G}$: 2 VG
    \item $\pi_{4VC}$: 2 VQA, 2 CLS
    \item $\pi_{4VCG}$: 2 VQA, 1 CLS, 1 VG
    \item $\pi_{8VC}$: 4 VQA, 4 CLS
    \item $\pi_{8VCG}$: 3 VQA, 3 CLS, 2 VG
    \item $\pi_{16VC}$: 8 VQA, 8 CLS
    \item $\pi_{32VCG}$: 10 VQA, 12 CLS, 10 VG
    \item $\pi_{64VCG}$: 20 VQA, 22 CLS, 22 VG
    \item $\pi_{128VCG}$: 42 VQA, 42 CLS, 44 VG
\end{itemize}

For each dataset, the images are equally duplicated as needed so that each case contains 128 examples, matching the batch size of 128 used during training.

\subsection{Prompt}

Since we do not “cold start” our model with fine-tuning on generated reasoning captions, prompting becomes the only means to elicit reasoning from the base model. As a result, the design of the system prompt is especially critical. However, we found that overly task-specific prompts can actually harm performance, as we will discuss in a later ablation study. For this reason, our system prompt remains straightforward and is aligned with the approach used in DeepSeek:

\begin{tcolorbox}[colback=gray!10, colframe=gray!60, boxrule=0.5pt]
\justifying
\noindent A conversation between User and Assistant. The user asks a question, and the Assistant solves it. The assistant first thinks about the reasoning process in the mind and then 
provides the user with the answer. The reasoning process and answer are enclosed within <reasoning> </reasoning> and <answer> </answer> tags, respectively, i.e., <reasoning> reasoning process here </reasoning><answer> answer here </answer>
\end{tcolorbox}

In addition to adjusting the system prompts, the question formats also need to be modified. The original VQA and CLS formats typically end with \texttt{… Answer the question using a single word or phrase.}, which often leads the base model to ignore prior instructions and simply provide a one-word answer. To address this, we update the questions to \texttt{… Make your chain of thought reasoning and then answer the question using a single word or phrase.} Similarly, for the VG tasks, the original prompt \texttt{Output the bounding box of the following object in the image…} is revised to \texttt{Make your chain of thought reasoning and then output the bounding box of the following object in the image.} With these changes, we observe that--even if only rarely--the base model now sometimes begins with reasoning before giving the final answer.

%% file: sec/4_experiments.tex
\section{Experiments}
\label{sec:exp}

\subsection{Experimental Setup}

During the reinforcement learning stage, GRPO is applied within the TRL (Transformer Reinforcement Learning) framework using batched processing. The base code from \cite{openr1multimodal} was forked from GitHub, and both the dataset and reward functions were adapted to fit our setup. Training was conducted on HPC clusters equipped with NVIDIA H100 GPUs, with the number of GPUs ranging from 16 to 128 depending on availability. Temperature is set to 0.9. Gradient accumulation steps were tuned between 1 and 8 to maintain a consistent overall batch size of 128. The Adam optimizer was used, with an initial learning rate of $1 \times 10^{-6}$. Due to memory constraints--even on H100 GPUs--we sampled 4 responses per image instead of the default 8. Following \cite{wang2025reinforcement} and our own ablation studies, we found that setting $\beta$ to 0.001 (rather than the default 0.04) yielded better results, so we adopted this value. Both the maximum prompt length and maximum completion length were set to 8192 tokens. The GRPO training phase was conducted for at least 1000 steps in each experiment, corresponding to 1000 epochs with training sets of 128 examples. If the performance continued to improve after 1000 steps, training was extended up to 2000 steps; otherwise, it was stopped.

\begin{figure*}[htbp]
    \centering
    \includegraphics[width=1\linewidth]{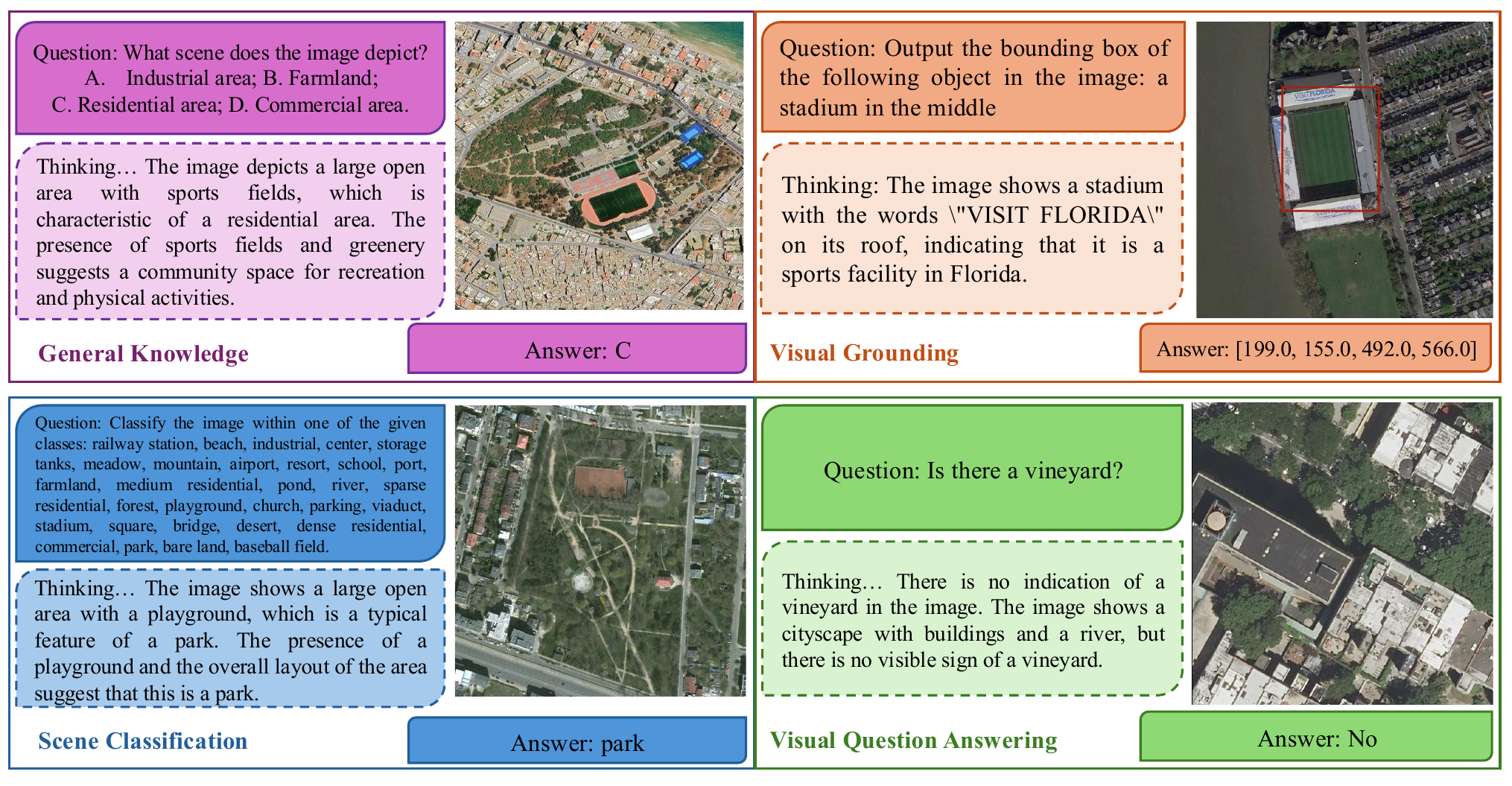}
    \caption{Qualitative outputs of our proposed method, trained with $\pi_{128VCG}$ dataset.}
    \label{fig:qualitative}
\end{figure*}

\subsection{Evaluation Setup}

We present our results on four popular remote sensing (RS) evaluation metrics: Remote Sensing Visual Question Answering (RSVQA), RS Classification, Visual Grounding, and General Knowledge (LHRS-Bench) \cite{muhtar2024lhrs, pang2025vhm}. For all evaluations, the temperature is set to 1. We evaluate all checkpoints of the trained models and report the ones that achieve the best results.

\subsection{Results}

The results of all trainings using the datasets mentioned from Section \ref{sec:dataset} is presented in Table \ref{tab:results}. Here are the key observations from the results:

\begin{itemize}
    \item Switching from the untouched base model to any of the 1-shot variants ($\pi_{1V}$, $\pi_{1C}$, $\pi_{1G}$) already delivers clear, double-digit boosts on most RS-Classification splits and sizeable gains on RSVQA and grounding. This validates our central hunch that verifiable-reward RL alone can unlock much of the latent reasoning ability without caption data.
    \item Training with a single example introduces issues related to the type of metric used for that specific case. For instance, using just one classification example for training negatively impacts the model’s classification performance, especially on METER-ML--the dataset from which the example is drawn. This suggests that training with only one example leads to some degree of overfitting, but this effect appears to be confined to evaluation examples of the same type; the model still generalizes well to other tasks.
    \item The 2-shot, 4-shot, and 8-shot cases do not achieve the best performance in any setting, but they consistently follow the baseline closely. They remain robust in most cases, and the “limited overfitting” observed in the 1-shot setting does not appear at all.
    \item Adding more examples pushes performance up across the board, yet the slope flattens after roughly 32–64 shots. By 128 shots the model is tracking the 2k image VHM-RL baseline in most categories, confirming diminishing returns and suggesting a pragmatic sweet spot around a few dozen curated cases.
    \item Estimating the pixel locations of a bounding box makes visual grounding by far the most challenging metric in our evaluation. It is therefore unsurprising that training with the full dataset outperforms the few-shot cases. Nonetheless, training with just 128 examples still yields promising results, suggesting that a few-shot approach focused specifically on VG data could further improve grounding performance to levels much closer to those achieved with full-data training.
    \item In all datasets, the training set lacks question-answer pairs similar to those in the LHRS-Bench evaluation set. As a result, the LHRS-Bench score serves as a measure of the model’s high level generalization ability. We observe that all models trained with 8-shot or more examples outperform VHM-RL on the LHRS-Bench metric. This demonstrates that few-shot training enables generalization, even when the training and evaluation sets are not similar.
\end{itemize}

Some qualitative results from the model trained on the $\pi_{128VCG}$ dataset are presented in Figure \ref{fig:qualitative}. These examples suggest that a brief reasoning process is sufficient for the model to answer this type of question effectively. Since the model was not specifically trained on reasoning data, it does not attempt to mimic human thinking patterns or use phrases like \textit{“oh wait, let me think”} and similar expressions. Thus, we can conclude that a concise chain-of-thought process is adequate for vision-language models to arrive at correct answers in these cases.

\subsection{Comparison with Other RS VLMs}

Although the primary aim of this paper is to analyze the effectiveness of few-shot training rather than introduce a new remote sensing language model, we include a comparison with existing algorithms trained on millions of examples. As shown in Table \ref{tab:results_other}, our models--trained on just 128 and 2000 images--achieve results that are competitive with these large-scale methods. Moreover, the existing models, except TinyRS \cite{koksal2025tinyrs}, are all 7B models, while our models have only 2B parameters. This highlights the remarkable potential of RLVR-based reasoning and training, even with a dramatically reduced amount of data.

\begin{table}[ht]
    \centering
    \caption{Performance comparison of our method with existing RS VLMs.}
    \begin{tabular}{l|c|c|c|c}
        \toprule
        Method & CLS & VQA & VG & Know. \\
        \midrule
        GeoChat \cite{kuckreja2024geochat} & 67.3 & \textbf{83.5} & 19.7 & 36.2 \\
        VHM \cite{pang2025vhm} & \textbf{85.6} & \underline{83.0}& 55.9 & 33.0\\
        TinyRS-R1 \cite{koksal2025tinyrs} & \textbf{85.6} & 76.0 & \textbf{74.9} & 56.8 \\
        Qwen2-VL-RS-R1 \cite{muhtar2025quality} & \underline{79.3} & 74.5 & \underline{64.6} & \textbf{69.2} \\
        VHM-RL-2000 \textbf{(Ours)} & 70.5 & 70.6 & 61.8 & 55.8 \\
        VHM-RL-128 \textbf{(Ours)} & 63.8 & 63.9 & 51.7 & \underline{60.0} \\
        \bottomrule
    \end{tabular}
    \label{tab:results_other}
\end{table}

%% file: sec/5_ablation.tex
\section{Ablation Study}
\label{sec:ablation}

\subsection{Effect of $\beta$}

In our main experiments, we follow \cite{wang2025reinforcement} and set the $\beta$ value for KL-divergence in the GRPO training algorithm to 0.001. For ablation purposes, we also experiment with the original default value of 0.04. In this study, we compare the training and evaluation results of models trained on the $\pi_{16VC}$ dataset under both $\beta$ settings.

\begin{table}[htbp]
    \centering
    \caption{Effect of $\beta$ on RLVR for the model trained on the $\pi_{16VC}$ dataset}
    \begin{tabular}{l|c|c|c|c}
        \toprule
        Method & CLS & VQA & VG & Know. \\
        \midrule
        $\beta$=0.001 & \textbf{56.3} & \textbf{65.8} & \textbf{31.4} & 57.2 \\
        $\beta$=0.04 & 53.9 & 63.6 & 30.7 & \textbf{57.8}\\
        \bottomrule
    \end{tabular}
    \label{tab:ab_beta}
\end{table}

\begin{figure}[htbp]
    \centering
    \includegraphics[width=\linewidth]{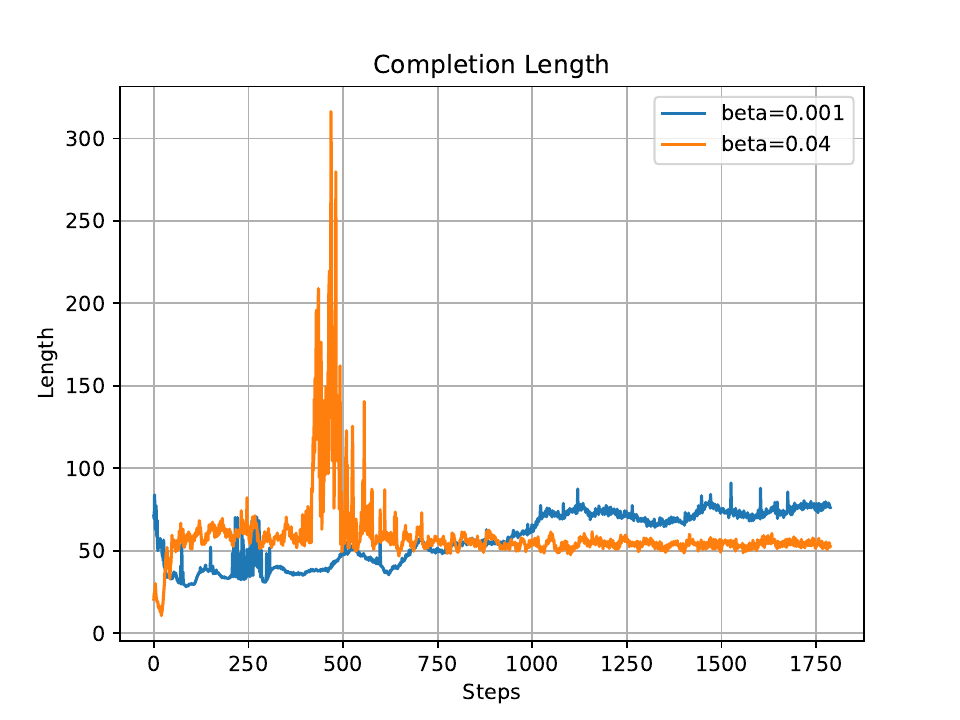}
    \caption{Change of completion length during training for the model trained on the $\pi_{16VC}$ dataset}
    \label{fig:completion}
\end{figure}

According to the results in Table \ref{tab:ab_beta}, the model trained with $\beta = 0.001$ outperforms the one trained with $\beta = 0.04$, though the improvement is incremental. Moreover, setting $\beta = 0.04$ introduces instability during training, which negatively impacts performance. As shown in Figure \ref{fig:completion}, the completion length for $\beta = 0.04$ increases abruptly for approximately 200 steps, causing the training process to become extremely slow before abruptly returning to normal. During this period, the model generates long, often nonsensical responses that do not contribute to learning. In contrast, training with $\beta = 0.001$ results in much more stable completion lengths and steadily improving response quality, as expected in a GRPO training.

\subsection{Effect of Prompting}

In this study, we examine the impact of both basic and detailed prompting. Intuitively, one might expect that, since the model has not been trained on any chain-of-thought data, providing a highly detailed prompt would be more effective. To test this hypothesis, we designed an alternative system prompt with extensive detail covering a wide range of possible scenarios, and retrained the model on the $\pi_{16VC}$ dataset using this prompt. The same detailed prompt was also used during evaluation:

\begin{tcolorbox}[colback=gray!10, colframe=gray!60, boxrule=0.5pt]
\justifying
\noindent You are an expert to analyze the image and provide useful information for users. Imitate a chain of thought reasoning process like a human brain with as detail as possible. Use expressions like oh wait, let me think etc. Then reach to the conclusion and provide an answer. If question starts with VQA with brackets, final answer needs be one word only as yes or no, rural or urban etc. If question stars with CLS, you need to answer with one of the given classes, nothing else. If question starts with VG then mentions an object, you need to locate the object on the image and give the output [[x\_min, y\_min, x\_max, y\_max]], nothing else. Keep the values between 0 and 1000. If none of the abbreviations with brackets are given at the beginning, you will answer the question with one or two sentences. The reasoning process and answer are enclosed within <reasoning> </reasoning> and <answer> </answer> tags, respectively, i.e., <reasoning>...your reasoning here...</reasoning> <answer>...your answer here...</answer>
\end{tcolorbox}

\begin{table}[htbp]
    \centering
    \caption{Effect of prompting on RLVR for the model trained on the $\pi_{16VC}$ dataset}    
    \begin{tabular}{l|c|c|c|c}
        \toprule
        Method & CLS & VQA & VG & Know. \\
        \midrule
        Basic & \textbf{56.3} & \textbf{65.8} & \textbf{31.4} & \textbf{57.2} \\
        Detailed & 38.2 & 64.3 & 17.9 & 52.4\\
        \bottomrule
    \end{tabular}
    \label{tab:ab_prompting}
\end{table}

According to the results in Table \ref{tab:ab_prompting}, performance actually declined with detailed prompting rather than improving. A likely explanation is that Qwen2-VL was initially pretrained with short system prompts and shorter token lengths. As a result, providing longer prompts during RLVR--without an accompanying pretraining phase--may not be effective, as the model is not conditioned to handle such input lengths.

\subsection{Effect of Model Size}

As previously discussed in Section \ref{sec:model}, the robustness of few-shot RLVR is independent of the underlying LLM architecture \cite{wang2025reinforcement}. However, it remains important to analyze the effect of model size within the same architecture. In this study, we compare the zero-shot Qwen2-VL-2B and Qwen2-VL-7B models with their few-shot RLVR-trained counterparts using the $\pi_{16VC}$ dataset.

\begin{table}[htbp]
    \centering
    \caption{Effect of model size on RLVR for the model trained on the $\pi_{16VC}$ dataset}
    \scalebox{0.9}{
    \begin{tabular}{c|c|c|c|c|c}
        \toprule
        Size & Train & CLS & VQA & VG & Know. \\
        \midrule
        2B & $\times$ & 48.8 & 33.1 & 9.5 & 52.6 \\
        2B & \checkmark & 56.3$\color{teal}^{+7.5}$ & 65.8$\color{teal}^{+34.7}$ & 31.4$\color{teal}^{+21.9}$ & 57.2$\color{teal}^{+4.6}$ \\
        \midrule
        7B & $\times$ & 63.0 & 61.2 & 14.3 & 57.0 \\
        7B & \checkmark & 62.0$\color{red}^{-1.0}$ & 68.7$\color{teal}^{+7.5}$ & 40.2$\color{teal}^{+25.9}$ & 67.2$\color{teal}^{+10.2}$ \\
        \bottomrule
    \end{tabular}}
    \label{tab:ab_model}
\end{table}

According to the results in Table \ref{tab:ab_model}, RLVR remains effective when applied to a larger-scale LLM with the same architecture, although the performance gains are somewhat limited. This is primarily because the 7B-parameter Qwen2-VL model is already substantially more capable on remote sensing tasks in zero-shot setting than its 2B counterpart. Nevertheless, the overall performance still improves significantly with only 16 training examples, demonstrating the robustness of RLVR across different-sized versions of the model.

%% file: sec/6_conclusion.tex
\section{Conclusion}
\label{sec:conc}

We present the first few-shot RLVR framework for vision–language reasoning in remote sensing, eliminating the need for costly caption supervision by relying solely on lightweight, rule-based rewards. Across four standard evaluation suites, even a single curated example attains double-digit gains over the base Qwen2-VL-2B, while scaling to 128 examples closes the gap to a 2000-image VHM-RL system and, on LHRS-Bench, surpasses it, affirming that minimal data combined with verifiable rewards can unlock the latent reasoning ability already present in compact VLMs.

Our analysis reveals three key insights. First, the well-publicised “1-shot RLVR” phenomenon generalises to multimodal satellite imagery, but the extreme setting can introduce localized overfitting--most evident when the lone example shares its metric with the test split--whereas 8–32 shots deliver stable, broadly transferable gains. Second, design choices matter: a concise system prompt and a small KL-weight ($\beta$ = 0.001) yield smoother optimisation and higher final accuracy than verbose prompting or the GRPO default $\beta$ = 0.04. Third, visual grounding remains the hardest sub-task; nonetheless, few-shot RLVR still lifts IoU@0.5 precision markedly, hinting that a grounding-focused shot allocation could shrink the remaining gap.

These findings suggest a pragmatic recipe for data-scarce domains: begin with an off-the-shelf small VLM, craft a handful of reward-checkable examples, and fine-tune with GRPO. The resulting model inherits the efficiency of its backbone while achieving specialist competence without large-scale annotation.

\textbf{Limitations \& future work.} A key limitation of our study is the mild, task-specific overfitting observed in the extreme one-shot scenario, suggesting that careful example selection remains crucial despite the general effectiveness of our approach. Additionally, although our random sampling method is effective, exploring more strategic selection methods, such as variance-based sampling, could further improve model performance and robustness. Future work will also investigate extending the few-shot RLVR approach to larger vision-language models and other specialized remote sensing tasks, such as temporal analysis or multi-sensor fusion, where expert annotation is even more costly. Moreover, refining the prompt engineering strategy to better align with the pretrained model characteristics could further enhance training stability and generalization.


\section*{Acknowledgements}

The numerical calculations reported in this paper were partially performed using the MareNostrum 5 pre-exascale supercomputing system and TÜBİTAK ULAKBİM, High Performance and Grid Computing Center (TRUBA resources). We gratefully acknowledge the Barcelona Supercomputing Center (BSC) and the Scientific and Technological Research Council of Turkey (TÜBİTAK) for providing access to these resources and supporting this research.

%% file: main.bbl
\begin{thebibliography}{46}
\providecommand{\natexlab}[1]{#1}
\providecommand{\url}[1]{\texttt{#1}}
\expandafter\ifx\csname urlstyle\endcsname\relax
  \providecommand{\doi}[1]{doi: #1}\else
  \providecommand{\doi}{doi: \begingroup \urlstyle{rm}\Url}\fi

\bibitem[Achiam et~al.(2023)Achiam, Adler, Agarwal, Ahmad, Akkaya, Aleman, Almeida, Altenschmidt, Altman, Anadkat, et~al.]{achiam2023gpt}
Josh Achiam, Steven Adler, Sandhini Agarwal, Lama Ahmad, Ilge Akkaya, Florencia~Leoni Aleman, Diogo Almeida, Janko Altenschmidt, Sam Altman, Shyamal Anadkat, et~al.
\newblock Gpt-4 technical report.
\newblock \emph{arXiv preprint arXiv:2303.08774}, 2023.

\bibitem[Agent(2025)]{openr1v}
Deep Agent.
\newblock Open-r1v.
\newblock https://github.com/Deep-Agent/R1-V, 2025.

\bibitem[Bai et~al.(2023{\natexlab{a}})Bai, Bai, Chu, Cui, Dang, Deng, Fan, Ge, Han, Huang, Hui, Ji, Li, Zhou, Zhou, Zhou, Zhu, et~al.]{bai2023qwentechnicalreport}
Jinze Bai, Shuai Bai, Yunfei Chu, Zeyu Cui, Kai Dang, Xiaodong Deng, Yang Fan, Wenbin Ge, Yu Han, Fei Huang, Binyuan Hui, Luo Ji, Mei Li, Chang Zhou, Jingren Zhou, Xiaohuan Zhou, Tianhang Zhu, et~al.
\newblock Qwen technical report, 2023{\natexlab{a}}.

\bibitem[Bai et~al.(2023{\natexlab{b}})Bai, Bai, Yang, Wang, Tan, Wang, Lin, Zhou, and Zhou]{bai2023qwen}
Jinze Bai, Shuai Bai, Shusheng Yang, Shijie Wang, Sinan Tan, Peng Wang, Junyang Lin, Chang Zhou, and Jingren Zhou.
\newblock Qwen-vl: A frontier large vision-language model with versatile abilities.
\newblock \emph{arXiv preprint arXiv:2308.12966}, 2023{\natexlab{b}}.

\bibitem[Bazi et~al.(2024)Bazi, Bashmal, Al~Rahhal, Ricci, and Melgani]{bazi2024rs}
Yakoub Bazi, Laila Bashmal, Mohamad~Mahmoud Al~Rahhal, Riccardo Ricci, and Farid Melgani.
\newblock Rs-llava: A large vision-language model for joint captioning and question answering in remote sensing imagery.
\newblock \emph{Remote Sensing}, 16\penalty0 (9):\penalty0 1477, 2024.

\bibitem[Beyer et~al.(2024)Beyer, Steiner, Pinto, Kolesnikov, Wang, Salz, Neumann, Alabdulmohsin, Tschannen, Bugliarello, et~al.]{beyer2024paligemma}
Lucas Beyer, Andreas Steiner, Andr{\'e}~Susano Pinto, Alexander Kolesnikov, Xiao Wang, Daniel Salz, Maxim Neumann, Ibrahim Alabdulmohsin, Michael Tschannen, Emanuele Bugliarello, et~al.
\newblock Paligemma: A versatile 3b vlm for transfer.
\newblock \emph{arXiv preprint arXiv:2407.07726}, 2024.

\bibitem[Brown et~al.(2020)Brown, Mann, Ryder, Subbiah, Kaplan, Dhariwal, Neelakantan, Shyam, Sastry, Askell, et~al.]{Brown2020}
Tom Brown, Benjamin Mann, Nick Ryder, Melanie Subbiah, Jared~D Kaplan, Prafulla Dhariwal, Arvind Neelakantan, Pranav Shyam, Girish Sastry, Amanda Askell, et~al.
\newblock Language models are few-shot learners.
\newblock \emph{Advances in neural information processing systems}, 33:\penalty0 1877--1901, 2020.

\bibitem[Bucher and Martini(2024)]{bucher2024fine}
Martin Juan~Jos{\'e} Bucher and Marco Martini.
\newblock Fine-tuned'small'llms (still) significantly outperform zero-shot generative ai models in text classification.
\newblock \emph{arXiv preprint arXiv:2406.08660}, 2024.

\bibitem[Chen et~al.(2024)Chen, Wu, Wang, Su, Chen, Xing, Zhong, Zhang, Zhu, Lu, et~al.]{chen2024internvl}
Zhe Chen, Jiannan Wu, Wenhai Wang, Weijie Su, Guo Chen, Sen Xing, Muyan Zhong, Qinglong Zhang, Xizhou Zhu, Lewei Lu, et~al.
\newblock Internvl: Scaling up vision foundation models and aligning for generic visual-linguistic tasks.
\newblock In \emph{Proceedings of the IEEE/CVF Conference on Computer Vision and Pattern Recognition}, pages 24185--24198, 2024.

\bibitem[Dehghani et~al.(2023)Dehghani, Mustafa, Djolonga, Heek, Minderer, Caron, Steiner, Puigcerver, Geirhos, Alabdulmohsin, et~al.]{dehghani2023patch}
Mostafa Dehghani, Basil Mustafa, Josip Djolonga, Jonathan Heek, Matthias Minderer, Mathilde Caron, Andreas Steiner, Joan Puigcerver, Robert Geirhos, Ibrahim~M Alabdulmohsin, et~al.
\newblock Patch n’pack: Navit, a vision transformer for any aspect ratio and resolution.
\newblock \emph{Advances in Neural Information Processing Systems}, 36:\penalty0 2252--2274, 2023.

\bibitem[Devlin et~al.(2018)Devlin, Chang, Lee, and Toutanova]{Devlin2019}
Jacob Devlin, Ming-Wei Chang, Kenton Lee, and Kristina Toutanova.
\newblock Bert: Pre-training of deep bidirectional transformers for language understanding.
\newblock \emph{arXiv preprint arXiv:1810.04805}, 2018.

\bibitem[Dosovitskiy et~al.(2020)Dosovitskiy, Beyer, Kolesnikov, Weissenborn, Zhai, Unterthiner, Dehghani, Minderer, Heigold, Gelly, et~al.]{dosovitskiy2020image}
Alexey Dosovitskiy, Lucas Beyer, Alexander Kolesnikov, Dirk Weissenborn, Xiaohua Zhai, Thomas Unterthiner, Mostafa Dehghani, Matthias Minderer, Georg Heigold, Sylvain Gelly, et~al.
\newblock An image is worth 16x16 words: Transformers for image recognition at scale.
\newblock \emph{arXiv preprint arXiv:2010.11929}, 2020.

\bibitem[Gandhi et~al.(2025)Gandhi, Chakravarthy, Singh, Lile, and Goodman]{gandhi2025cognitive}
Kanishk Gandhi, Ayush Chakravarthy, Anikait Singh, Nathan Lile, and Noah~D Goodman.
\newblock Cognitive behaviors that enable self-improving reasoners, or, four habits of highly effective stars.
\newblock \emph{arXiv preprint arXiv:2503.01307}, 2025.

\bibitem[Guo et~al.(2025)Guo, Yang, Zhang, Song, Zhang, Xu, Zhu, Ma, Wang, Bi, et~al.]{guo2025deepseek}
Daya Guo, Dejian Yang, Haowei Zhang, Junxiao Song, Ruoyu Zhang, Runxin Xu, Qihao Zhu, Shirong Ma, Peiyi Wang, Xiao Bi, et~al.
\newblock Deepseek-r1: Incentivizing reasoning capability in llms via reinforcement learning.
\newblock \emph{arXiv preprint arXiv:2501.12948}, 2025.

\bibitem[Hu et~al.(2022)Hu, Shen, Wallis, Allen-Zhu, Li, Wang, and Chen]{Hu2022}
Edward Hu, Yining Shen, Peter Wallis, Zeyuan Allen-Zhu, Yuanzhi Li, Lijuan Wang, and Weizhu Chen.
\newblock Lora: Low-rank adaptation of large language models.
\newblock \emph{arXiv preprint arXiv:2106.09685}, 2022.

\bibitem[Hu et~al.(2023)Hu, Yuan, Wen, Lu, and Li]{hu2307remote}
Y Hu, J Yuan, C Wen, X Lu, and X Li.
\newblock A remote sensing vision language model and benchmark. arxiv 2023.
\newblock \emph{arXiv preprint arXiv:2307.15266}, 2023.

\bibitem[Huggingface(2025)]{openr1}
Huggingface.
\newblock Open-r1.
\newblock https://github.com/huggingface/open-r1, 2025.

\bibitem[Hurst et~al.(2024)Hurst, Lerer, Goucher, Perelman, Ramesh, Clark, Ostrow, Welihinda, Hayes, Radford, et~al.]{hurst2024gpt}
Aaron Hurst, Adam Lerer, Adam~P Goucher, Adam Perelman, Aditya Ramesh, Aidan Clark, AJ Ostrow, Akila Welihinda, Alan Hayes, Alec Radford, et~al.
\newblock Gpt-4o system card.
\newblock \emph{arXiv preprint arXiv:2410.21276}, 2024.

\bibitem[Koksal and Alatan(2025{\natexlab{a}})]{koksal2025milchat}
Aybora Koksal and A~Aydin Alatan.
\newblock Milchat: Introducing chain of thought reasoning and grpo to a multimodal small language model for remote sensing.
\newblock \emph{arXiv preprint arXiv:2505.07984}, 2025{\natexlab{a}}.

\bibitem[Koksal and Alatan(2025{\natexlab{b}})]{koksal2025tinyrs}
Aybora Koksal and A~Aydin Alatan.
\newblock Tinyrs-r1: Compact multimodal language model for remote sensing.
\newblock \emph{arXiv preprint arXiv:2505.12099}, 2025{\natexlab{b}}.

\bibitem[Kuckreja et~al.(2024)Kuckreja, Danish, Naseer, Das, Khan, and Khan]{kuckreja2024geochat}
Kartik Kuckreja, Muhammad~Sohail Danish, Muzammal Naseer, Abhijit Das, Salman Khan, and Fahad~Shahbaz Khan.
\newblock Geochat: Grounded large vision-language model for remote sensing.
\newblock In \emph{Proceedings of the IEEE/CVF Conference on Computer Vision and Pattern Recognition}, pages 27831--27840, 2024.

\bibitem[Li et~al.(2024)Li, Muhtar, Gu, Zhang, Xiao, He, and Zhu]{li2024lhrs}
Zhenshi Li, Dilxat Muhtar, Feng Gu, Xueliang Zhang, Pengfeng Xiao, Guangjun He, and Xiaoxiang Zhu.
\newblock Lhrs-bot-nova: Improved multimodal large language model for remote sensing vision-language interpretation.
\newblock \emph{arXiv preprint arXiv:2411.09301}, 2024.

\bibitem[Liu et~al.(2023)Liu, Li, Wu, and Lee]{liu2023llava}
Haotian Liu, Chunyuan Li, Qingyang Wu, and Yong~Jae Lee.
\newblock Visual instruction tuning, 2023.

\bibitem[Liu et~al.(2025)Liu, Chen, Li, Qi, Pang, Du, Lee, and Lin]{liu2025understanding}
Zichen Liu, Changyu Chen, Wenjun Li, Penghui Qi, Tianyu Pang, Chao Du, Wee~Sun Lee, and Min Lin.
\newblock Understanding r1-zero-like training: A critical perspective.
\newblock \emph{arXiv preprint arXiv:2503.20783}, 2025.

\bibitem[LMMs-Lab(2025)]{openr1multimodal}
LMMs-Lab.
\newblock Open-r1 multimodal.
\newblock https://github.com/EvolvingLMMs-Lab/open-r1-multimodal, 2025.

\bibitem[Muhtar et~al.(2024)Muhtar, Li, Gu, Zhang, and Xiao]{muhtar2024lhrs}
Dilxat Muhtar, Zhenshi Li, Feng Gu, Xueliang Zhang, and Pengfeng Xiao.
\newblock Lhrs-bot: Empowering remote sensing with vgi-enhanced large multimodal language model.
\newblock In \emph{European Conference on Computer Vision}, pages 440--457. Springer, 2024.

\bibitem[Muhtar et~al.(2025)Muhtar, Zhang, Li, Gu, He, Xiao, and Zhang]{muhtar2025quality}
Dilxat Muhtar, Enzhuo Zhang, Zhenshi Li, Feng Gu, Yanglangxing He, Pengfeng Xiao, and Xueliang Zhang.
\newblock Quality-driven curation of remote sensing vision-language data via learned scoring models.
\newblock \emph{arXiv preprint arXiv:2503.00743}, 2025.

\bibitem[OpenAI(2023{\natexlab{a}})]{gpt35turbo}
OpenAI.
\newblock Gpt-3.5 turbo fine-tuning and api updates, 2023{\natexlab{a}}.

\bibitem[OpenAI(2023{\natexlab{b}})]{gpt4v}
OpenAI.
\newblock Gpt-4v(ision) system card, 2023{\natexlab{b}}.

\bibitem[OpenAI(2024)]{o1preview}
OpenAI.
\newblock Introducing openai o1-preview, 2024.

\bibitem[OpenAI(2025)]{o3mini}
OpenAI.
\newblock Openai o3-mini, 2025.

\bibitem[Pang et~al.(2025)Pang, Weng, Wu, Li, Liu, Sun, Li, Wang, Feng, Xia, et~al.]{pang2025vhm}
Chao Pang, Xingxing Weng, Jiang Wu, Jiayu Li, Yi Liu, Jiaxing Sun, Weijia Li, Shuai Wang, Litong Feng, Gui-Song Xia, et~al.
\newblock Vhm: Versatile and honest vision language model for remote sensing image analysis.
\newblock In \emph{Proceedings of the AAAI Conference on Artificial Intelligence}, pages 6381--6388, 2025.

\bibitem[Radford et~al.(2018)Radford, Narasimhan, Salimans, Sutskever, et~al.]{radford2018improving}
Alec Radford, Karthik Narasimhan, Tim Salimans, Ilya Sutskever, et~al.
\newblock Improving language understanding by generative pre-training.
\newblock \emph{Technical Report}, 2018.

\bibitem[Radford et~al.(2021)Radford, Kim, Hallacy, Ramesh, Goh, Agarwal, Sastry, Askell, Mishkin, Clark, et~al.]{radford2021learning}
Alec Radford, Jong~Wook Kim, Chris Hallacy, Aditya Ramesh, Gabriel Goh, Sandhini Agarwal, Girish Sastry, Amanda Askell, Pamela Mishkin, Jack Clark, et~al.
\newblock Learning transferable visual models from natural language supervision.
\newblock \emph{arXiv preprint arXiv:2103.00020}, 2021.

\bibitem[Shah et~al.(2025)Shah, Rushton, Singla, Parmar, Smith, Vanjani, Vaswani, Chaluvaraju, Hojel, Ma, et~al.]{shah2025rethinking}
Darsh~J Shah, Peter Rushton, Somanshu Singla, Mohit Parmar, Kurt Smith, Yash Vanjani, Ashish Vaswani, Adarsh Chaluvaraju, Andrew Hojel, Andrew Ma, et~al.
\newblock Rethinking reflection in pre-training.
\newblock \emph{arXiv preprint arXiv:2504.04022}, 2025.

\bibitem[Shao et~al.(2024)Shao, Wang, Zhu, Xu, Song, Bi, Zhang, Zhang, Li, Wu, et~al.]{shao2024deepseekmath}
Zhihong Shao, Peiyi Wang, Qihao Zhu, Runxin Xu, Junxiao Song, Xiao Bi, Haowei Zhang, Mingchuan Zhang, YK Li, Y Wu, et~al.
\newblock Deepseekmath: Pushing the limits of mathematical reasoning in open language models.
\newblock \emph{arXiv preprint arXiv:2402.03300}, 2024.

\bibitem[Steiner et~al.(2024)Steiner, Pinto, Tschannen, Keysers, Wang, Bitton, Gritsenko, Minderer, Sherbondy, Long, et~al.]{steiner2024paligemma}
Andreas Steiner, Andr{\'e}~Susano Pinto, Michael Tschannen, Daniel Keysers, Xiao Wang, Yonatan Bitton, Alexey Gritsenko, Matthias Minderer, Anthony Sherbondy, Shangbang Long, et~al.
\newblock Paligemma 2: A family of versatile vlms for transfer.
\newblock \emph{arXiv preprint arXiv:2412.03555}, 2024.

\bibitem[Team et~al.(2024)Team, Mesnard, Hardin, Dadashi, Bhupatiraju, Pathak, Sifre, Rivière, Kale, Fiedel, Senter, Andreev, Kenealy, et~al.]{gemmateam2024gemmaopenmodelsbased}
Gemma Team, Thomas Mesnard, Cassidy Hardin, Robert Dadashi, Surya Bhupatiraju, Shreya Pathak, Laurent Sifre, Morgane Rivière, Mihir~Sanjay Kale, Noah Fiedel, Evan Senter, Alek Andreev, Kathleen Kenealy, et~al.
\newblock Gemma: Open models based on gemini research and technology, 2024.

\bibitem[Team(2023)]{2023internlm}
InternLM Team.
\newblock Internlm: A multilingual language model with progressively enhanced capabilities.
\newblock \url{https://github.com/InternLM/InternLM-techreport}, 2023.

\bibitem[Team et~al.(2025)Team, Du, Gao, Xing, Jiang, Chen, Li, Xiao, Du, Liao, et~al.]{team2025kimi}
Kimi Team, Angang Du, Bofei Gao, Bowei Xing, Changjiu Jiang, Cheng Chen, Cheng Li, Chenjun Xiao, Chenzhuang Du, Chonghua Liao, et~al.
\newblock Kimi k1. 5: Scaling reinforcement learning with llms.
\newblock \emph{arXiv preprint arXiv:2501.12599}, 2025.

\bibitem[Touvron et~al.(2023)Touvron, Lavril, Izacard, Martinet, Lachaux, Lacroix, Rozière, Goyal, Hambro, Azhar, Rodriguez, Joulin, Grave, and Lample]{touvron2023llama}
Hugo Touvron, Thibaut Lavril, Gautier Izacard, Xavier Martinet, Marie-Anne Lachaux, Timothée Lacroix, Baptiste Rozière, Naman Goyal, Eric Hambro, Faisal Azhar, Aurelien Rodriguez, Armand Joulin, Edouard Grave, and Guillaume Lample.
\newblock Llama: Open and efficient foundation language models, 2023.

\bibitem[Vaswani et~al.(2017)Vaswani, Shazeer, Parmar, Uszkoreit, Jones, Gomez, Kaiser, and Polosukhin]{Vaswani2017}
Ashish Vaswani, Noam Shazeer, Niki Parmar, Jakob Uszkoreit, Llion Jones, Aidan~N Gomez, Lukasz Kaiser, and Illia Polosukhin.
\newblock Attention is all you need.
\newblock In \emph{Advances in Neural Information Processing Systems}, pages 5998--6008, 2017.

\bibitem[Wang et~al.(2024)Wang, Bai, Tan, Wang, Fan, Bai, Chen, Liu, Wang, Ge, et~al.]{wang2024qwen2}
Peng Wang, Shuai Bai, Sinan Tan, Shijie Wang, Zhihao Fan, Jinze Bai, Keqin Chen, Xuejing Liu, Jialin Wang, Wenbin Ge, et~al.
\newblock Qwen2-vl: Enhancing vision-language model's perception of the world at any resolution.
\newblock \emph{arXiv preprint arXiv:2409.12191}, 2024.

\bibitem[Wang et~al.(2025)Wang, Yang, Zeng, Ren, Liu, Peng, Cheng, He, Wang, Gao, et~al.]{wang2025reinforcement}
Yiping Wang, Qing Yang, Zhiyuan Zeng, Liliang Ren, Liyuan Liu, Baolin Peng, Hao Cheng, Xuehai He, Kuan Wang, Jianfeng Gao, et~al.
\newblock Reinforcement learning for reasoning in large language models with one training example.
\newblock \emph{arXiv preprint arXiv:2504.20571}, 2025.

\bibitem[Yue et~al.(2025)Yue, Chen, Lu, Zhao, Wang, Song, and Huang]{yue2025does}
Yang Yue, Zhiqi Chen, Rui Lu, Andrew Zhao, Zhaokai Wang, Shiji Song, and Gao Huang.
\newblock Does reinforcement learning really incentivize reasoning capacity in llms beyond the base model?
\newblock \emph{arXiv preprint arXiv:2504.13837}, 2025.

\bibitem[Zhang et~al.(2024)Zhang, Cai, Zhang, Zhuang, and Mao]{zhang2024earthgpt}
Wei Zhang, Miaoxin Cai, Tong Zhang, Yin Zhuang, and Xuerui Mao.
\newblock Earthgpt: A universal multi-modal large language model for multi-sensor image comprehension in remote sensing domain.
\newblock \emph{IEEE Transactions on Geoscience and Remote Sensing}, 2024.

\end{thebibliography}
